\documentclass{isprs} 
\usepackage{setspace}
\usepackage{amsmath} 
\usepackage{booktabs}
\usepackage[table]{xcolor}
\definecolor{bestrow}{HTML}{FFF6BF} 
\usepackage{tabularx}
\usepackage{graphicx}
\usepackage{subcaption}
\usepackage{graphbox}
\usepackage{float}      % for [H]
\graphicspath{{figures/}} 
\usepackage{multirow}
\usepackage{algorithm}
\usepackage{algpseudocode}
\usepackage{adjustbox}  
\usepackage{makecell}  

\usepackage{geometry}
\usepackage[
  pdftex,
  hidelinks,
  hypertexnames=false
]{hyperref}
\usepackage{epstopdf}
\usepackage[labelsep=period]{caption}  
\usepackage[british]{babel} 
\usepackage[hang]{footmisc}
\usepackage{}

\usepackage{tikz}
\usetikzlibrary{positioning}
\usetikzlibrary{calc}
\usetikzlibrary{arrows.meta}
\usetikzlibrary{shapes.misc}
\usetikzlibrary{decorations.pathmorphing}
\usetikzlibrary{shapes.multipart}
\usetikzlibrary{fit}
\usetikzlibrary{backgrounds}

\usepackage[inline]{enumitem}
\usepackage{xspace}
\usepackage{siunitx}
\DeclareSIUnit[
  product-units=single,
  multi-part-units = single,
]\pixel{px}
\usepackage{csquotes}
\usepackage{amsfonts}
\usepackage{relsize}
\usepackage{microtype}
\usepackage{tumabbrev}
\usepackage{tumcolors}
\usepackage{tummath}
\usepackage{zref-clever}
\usepackage[useregional]{datetime2}

\zcsetup{cap,abbrev=false}

\usepackage[
  backend=biber,
  style=authoryear,
  natbib=true,
  giveninits=true,
  sorting=nyt,
  backref=false,
  dashed=false,
  alldates=year,
  urldate=long,
  maxcitenames=2,
  mincitenames=1,
  maxbibnames=3,
  url=true,
  doi=true,
  eprint=false,
  uniquename=false,
  uniquelist=false,
  isbn=false,
  firstinits=true,
]{biblatex}

\DeclareNameAlias{sortname}{family-given}
\AtBeginBibliography{\smaller[1]}
\DeclareSourcemap{
  \maps[datatype=bibtex, overwrite]{
    \map{
      \step[fieldset=editor, null]
    }
  }
}
\begin{document}

\title{deSEO: Physics-Aware Dataset Creation for High-Resolution Satellite Image Shadow Removal\thanks{Accepted in the annals track at the ISPRS 2026 Congress.}}
\date{}

\author{
  Lorenzo Beltrame\textsuperscript{1,4}, Jules Salzinger\textsuperscript{1}, Filip Svoboda\textsuperscript{2}, Phillipp Fanta-Jende\textsuperscript{1}, Jasmin Lampert\textsuperscript{1}, Radu Timofte\textsuperscript{3}, Marco Körner\textsuperscript{4,5,6}  }

\address{
    \textsuperscript{1 }Austrian Institute of Technology, Giefinggasse 4, 1210 Vienna, Austria -- (name.surname)@ait.ac.at\\
    \textsuperscript{2 }University of Cambridge, William Gates Building, 15 JJ Thomson Ave., CB3 0FD Cambridge, UK -- fs437@cam.ac.uk\\
    \textsuperscript{3 }University of W\"urzburg, John Skilton Str.\ 4a, Hubland Nord, 97074 W\"urzburg, Germany -- radu.timofte@uni-wuerzburg.de\\
    \textsuperscript{4 }Technical University of Munich (TUM), TUM School of Engineering and Design, Chair of Remote Sensing Technology,\\ Arcisstr. 21, 80333 Munich, Germany -- marco.koerner@tum.de\\
    \textsuperscript{5 }Technical University of Munich (TUM), Munich Data Science Institute (MDSI), 85748 Garching, Germany\\
    \textsuperscript{6 }ELLIS Unit Jena, Friedrich Schiller University of Jena, 07743 Jena, Germany\\
}

% information on the corresponding author should not be used any longer and has been commen ted out
% C. Heipke, Jan 03,2024

% the use of the information of commissions and working groups should not be used any longer and has been removed using comm ents out
% C. Heipke, Sept. 20,2022
%\commission{XX, }{YY} %This field is optional. If filled, XX and YY should be replaced by adequate numbers. See https://www2.isprs.org/commissions/
%\workinggroup{XX/YY} %This field is optional.
%\icwg{}   %This field is optional.

\abstract{
Shadows cast by terrain and tall structures remain a major obstacle for high-resolution satellite image analysis, degrading classification, detection, and 3D reconstruction performance. 
Public resources offering geometry-consistent paired shadow/shadow-free satellite imagery are essentially missing, and most Earth-observation datasets are designed for shadow detection or 3D modelling rather than removal.
Existing deep shadow-removal datasets either target ground-level or aerial scenes or rely on unpaired and weakly supervised formulations rather than explicit satellite pairs. 
We address this gap with \emph{deSEO}, a \emph{geometry-aware} and \emph{physics-informed} methodology that, to the best of our knowledge, is the first to derive paired supervision for satellite shadow removal from the S-EO shadow detection dataset \parencite{masquil2025shadoweo} through a fully replicable pipeline. 
For each tile, deSEO selects a minimally shadowed acquisition as a weak reference and pairs it with shadowed counterparts using temporal and geometric filtering, Jacobian-based orientation normalisation, and LoFTR–RANSAC registration. 
A per-pixel validity mask restricts learning to reliably aligned regions, enabling supervision despite residual off-nadir parallax. 
In addition to this paired dataset, we develop a DSM-aware deshadowing model that combines residual translation, perceptual objectives, and mask-constrained adversarial learning. 
In contrast, a direct adaptation of a UAV-based SRNet/pix2pix architecture fails to converge under satellite viewpoint variability.
Our model consistently reduces the visual impact of cast shadows across diverse illumination and viewing conditions, achieving improved structural and perceptual fidelity on held-out scenes. 
deSEO therefore provides the first reproducible, geometry-aware paired dataset and baseline for shadow removal in satellite Earth observation. 
}


\maketitle

\section{Introduction}\label{SECT: Introduction}

The abundance of \emph{Earth Observation (EO)} data in combination with advanced machine learning techniques is currently revolutionising the field of remote sensing. 
Among the many physical challenges to aerial and satellite-based remote sensing, terrain-induced shadows constitute an impediment to various forms of analysis \citep{li2016general}. 
This is particularly true in high-slope environments such as mountains \citep{giles2001remote}, but also in cities where man-made structures can cast strong shadows on other regions of interest \citep{dare2005shadow}. 
Physics-based methods, grounded in light transport theory, have long provided solutions for shadow detection and removal. 
Still, those methods are known to produce artifacts that can interfere with downstream processing \citep{le2021physics}. 
Following recent trends in various fields, deep learning-based methods have been successfully applied to this task \citep{dong2024review}, yielding improvements in both deshadowing and downstream task performance \citep{Zhu2024,Zhang2025}.

Existing shadow-aware EO datasets, such as the \emph{Aerial Imagery Shadow Detection (AISD)} dataset \citep{luo2020deeply} and the \emph{CUHK-Shadow} dataset \citep{hu2021revisiting}, are primarily designed for shadow detection, rather than their removal. 
They rely on manual annotations, which are costly and potentially subjective, and they do not provide true shadow-free references. 
While the large-scale \emph{Shadow-aware Earth Observation (S-EO)} dataset \citep{masquil2025shadoweo} advances detection with geometry-aware, automatically generated masks, it remains detection-oriented and does not provide paired acquisitions suitable for weakly supervised deshadowing. 
More broadly, current resources seldom control for seasonal shifts, viewing geometry differences, or illumination changes, all of which introduce confounding factors that hinder the training and fair evaluation of shadow-removal models in high-resolution satellite imagery.

A further challenge is that, despite recent progress in deep learning for shadow removal, there is currently no public available dataset of geometry-consistent, paired shadowed and shadow-free EO satellite imagery suitable for weakly supervised deshadowing. 
Existing shadow-removal datasets either focus on ground-level scenes \citep{qu2017deshadownet,wang2018STCGAN, vasluianu2023wsrd}, provide paired imagery acquired by \emph{unmanned aerial vehicles (UAVs)} under controlled conditions \citep{luo2023evolutionary}, or adopt unpaired and weakly supervised formulations without true shadow-free references \citep{wang2024recreating}. 
None of these resources captures the multi-date, multi-angle satellite imaging geometry, with \emph{rational polynomial coefficient (RPC)} camera models and \emph{digital surface model (DSM)} priors, that is required to build reliable cross-view shadow correspondences. 
Consequently, shadow-removal methods developed for close-range or UAV imagery do not transfer to satellite imagery, where parallax, seasonal variability, and radiometric inconsistencies dominate. This motivates the need for a methodology that can transform detection-oriented EO datasets into paired resources for deshadowing through principled geometric filtering and reproducible data processing.

To address these limitations, we introduce \emph{deSEO}, a data processing methodology that leverages multi-temporal views of a scene to translate shadow-detection datasets into shadow-removal datasets. 
Minimally shadowed images are selected as proxy references and paired with shadowed counterparts, enabling weakly supervised training despite the technical impossibility of perfectly shadow-free ground truth. 
The pairing enforces weak supervision under explicit constraints (\eg seasonal proximity, footprint overlap, and view-geometry similarity) to reduce bias from seasonal and illumination variations while preserving sufficient shadow contrast for learning. 
In doing so, deSEO reframes existing detection-focused resources into datasets engineered for weakly supervised deshadowing in high-resolution EO imagery.

Our contributions are threefold, \ie
\begin{enumerate}[label=\roman*), nosep]
    \item a \emph{geometry-aware pipeline} for constructing paired training samples for satellite deshadowing from a multi-acquisition EO shadow detection dataset through two-stage filtering, orientation normalisation, feature-based registration, and validity-aware pairing,
    \item the first paired, \emph{geometry-consistent} dataset for high resolution satellite shadow removal, derived from S-EO using the deSEO pipeline, as well as
    \item a \emph{deshadowing model} and \emph{training strategy} tailored to high-resolution satellite imagery, which uses registration-driven validity masks to restrict supervision to reliable correspondences and remain robust to residual misalignment.
\end{enumerate}

\section{Methodology}
\begin{figure*}[t]
    \centering
    \resizebox{0.85\textwidth}{!}{\input{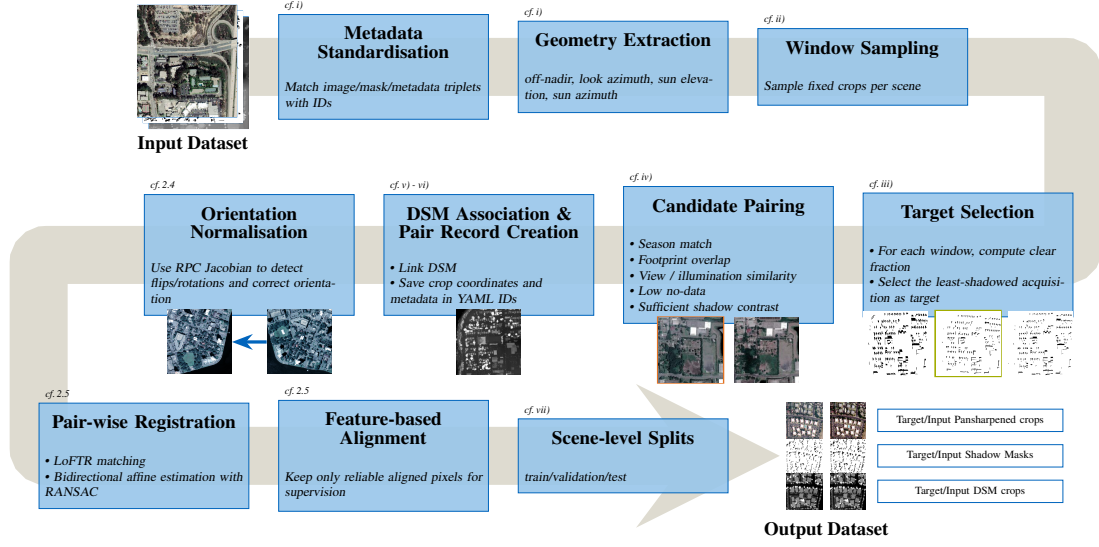}}
    \caption{Diagram of deSEO, the proposed framework for shadow removal dataset creation. The processing pipeline uses multitemporal acquisitions paired with a digital surface model and the relative shadow masks as input and outputs a machine-learning-ready dataset.}
    \label{fig:diagram}
\end{figure*}

We propose a preprocessing and training pipeline that makes the S-EO dataset~\citep{masquil2025shadoweo} suitable for weakly supervised deep learning of shadow removal in high-resolution satellite imagery.
The pipeline leverages S-EO's geometry-aware design and multi-temporal, multi-angle acquisitions to create paired samples that support data-driven deshadowing under realistic acquisition variability.
In the second part of this study, we propose a single-stage, weakly supervised network to perform deshadowing, leveraging the physical priors incorporated into S-EO.
A diagram of the shadow removal dataset creation deSEO is presented in \zcref{fig:diagram}.

\subsection{Dataset Overview: S-EO}
The S-EO dataset consists of \num{20000} georeferenced WorldView-3 images covering \num{702} tiles of \qtyproduct{500x500}{\meter} across the three cities of San Diego (UCSD), Omaha (OMA), and Jacksonville (JAX), acquired over several years to provide diverse solar and viewing geometries. 
Available modalities include 
\begin{itemize*}[label={}, itemjoin={, }, itemjoin*={, and }, after={.}]
    \item pan-sharpened RGB and panchromatic imagery (\qty[per-mode=symbol]{30}{\centi\meter\per\pixel})
    \item LiDAR-derived minimal and maximal DSM heights (\qty[per-mode=symbol]{50}{\centi\meter\per\pixel})
    \item physically generated shadow masks, NDVI vegetation masks
    \item bundle-adjusted RPC camera models
\end{itemize*}
These study areas span diverse urban and suburban morphologies, enabling evaluation of geographic generalisation.  
While the S-EO dataset provides both physically simulated shadow masks and corresponding uncertainty maps identifying unprojected or geometrically unreliable pixels, the latter were not used in the deSEO pairing process. 
This choice follows from the fundamental difference in supervision design: our goal is to construct paired, image-level correspondences rather than to refine pixel-wise shadow annotations. 
Because shadow detection is not the target task, incorporating the uncertainty masks would have excluded large regions of otherwise valid imagery, reducing the diversity of spatial and radiometric contexts available for pairing. 
Instead, deSEO introduces registration-driven validity masks that are derived directly from the multi-temporal alignment stage. 
These masks delimit reliable correspondences after geometric harmonisation, ensuring that losses are applied only where cross-view registration is confident, thereby serving the same purpose as the original uncertainty maps but in the context of weakly supervised deshadowing. 
\zcref[S]{fig:ucsd-dsm-shadow-rgb} shows an example of a pansharpened RGB image, a max shadow mask, and a DSM required for the deSEO pipeline.  
In the remainder, the shadow masks generated from the DSM are referred to as the \emph{shadow masks} for short.

\begin{figure}[t]
  \centering
  \subcaptionbox{DSM\label{fig:ucsd-dsm}}{\includegraphics[width=.32\linewidth,height=.32\linewidth]{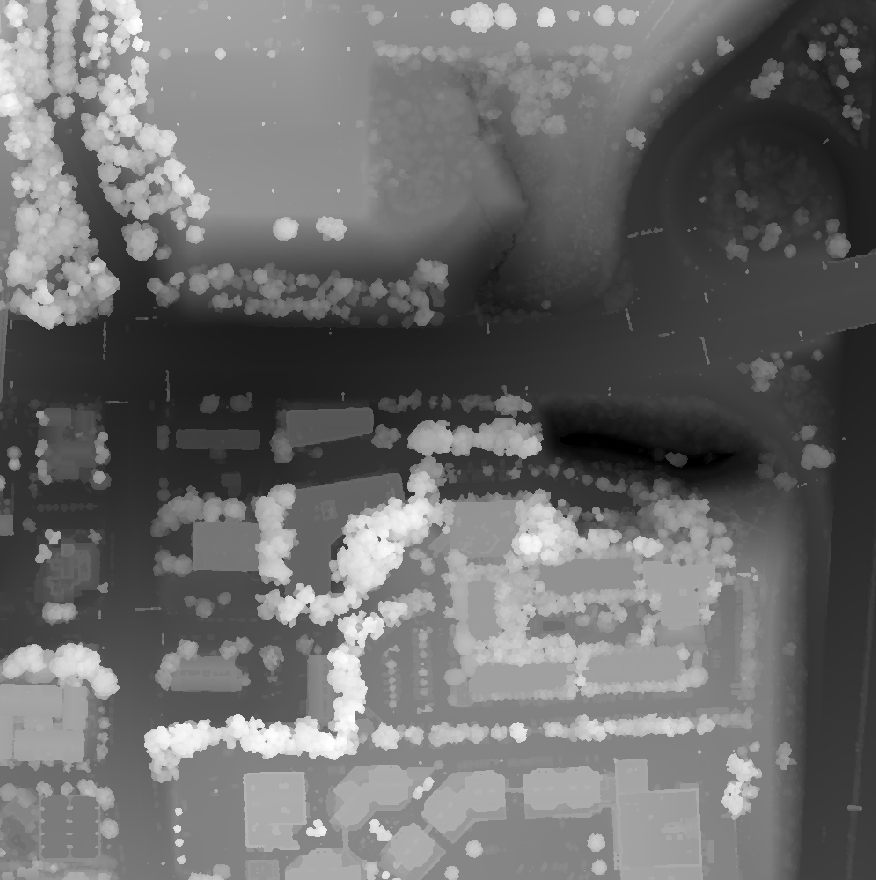}}\hfill
  \subcaptionbox{shadow mask\label{fig:ucsd-shadow}}{\includegraphics[width=.32\linewidth,height=.32\linewidth]{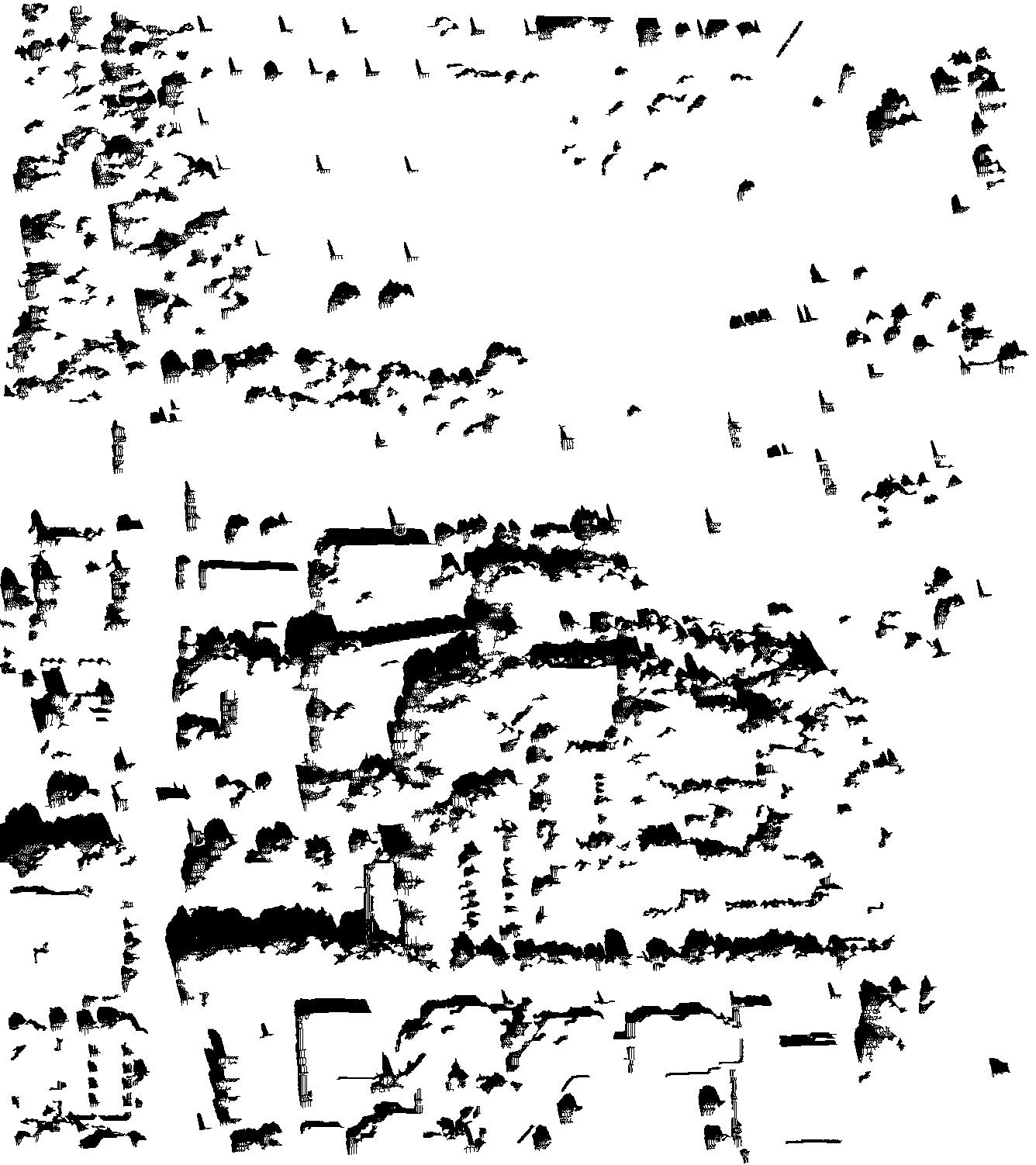}}\hfill
  \subcaptionbox{pan-sharpened RGB\label{fig:ucsd-rgb}}{\includegraphics[width=.32\linewidth,height=.32\linewidth]{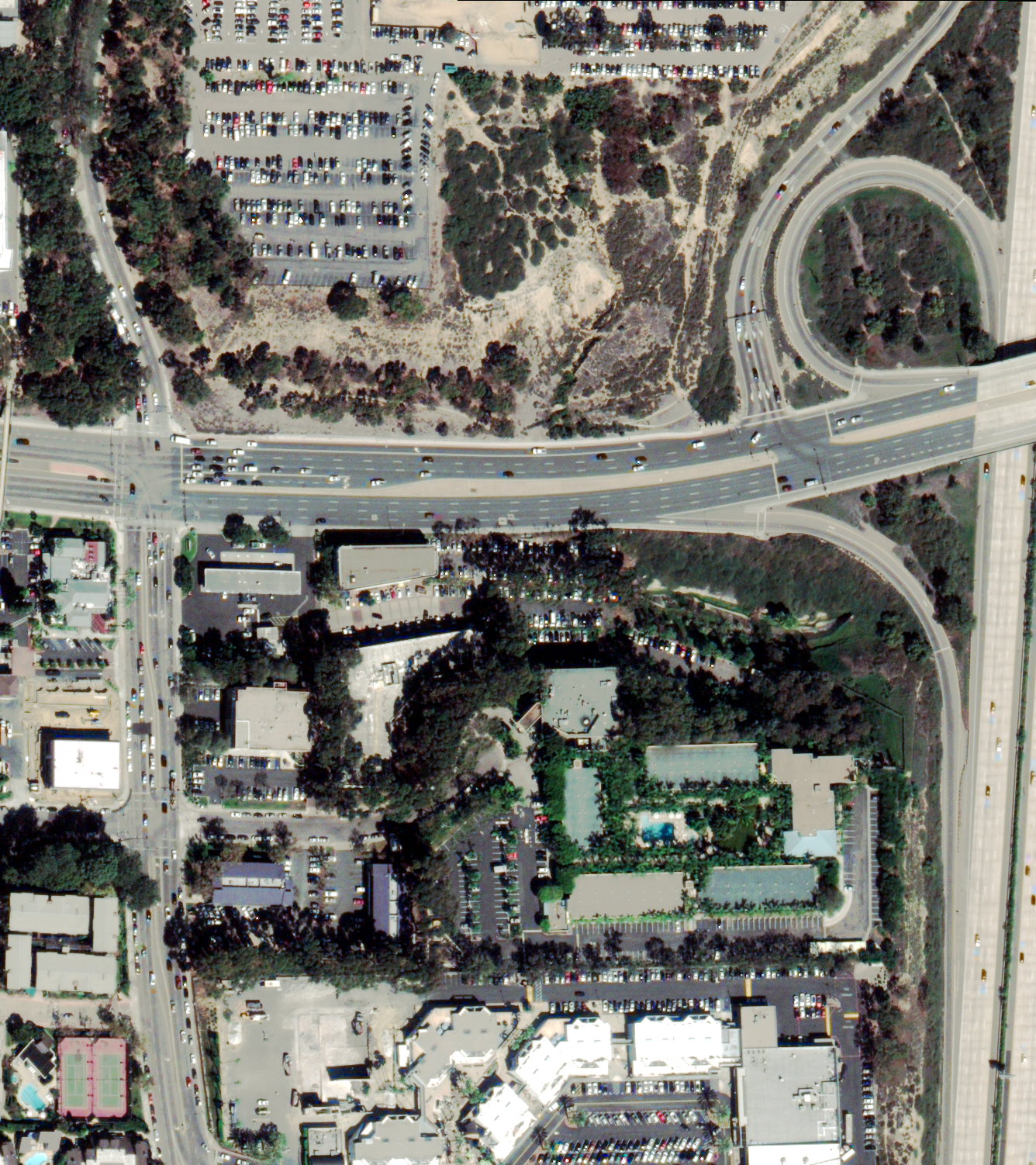}}
\caption{%
  Acquisition from the UCSD tile: 
  \textup{(a)} LiDAR-derived DSM (maximum aggregation), 
  \textup{(b)} the physically simulated shadow mask, and 
  \textup{(c)} the corresponding WorldView-3 RGB image. 
  The DSM footprint is larger than the RGB and mask because it is designed to cover all the offsets in the time series.}
  \label{fig:ucsd-dsm-shadow-rgb}
  % \vspace{-6pt}
\end{figure}

\subsection{Deshadowing Dataset Creation: deSEO}
It is possible to obtain a shadow-removal dataset from a shadow-detection dataset by exploiting paired crops from multi-temporal, high-resolution satellite imagery that contains a shadowy scene and a counterpart presenting less pronounced shadows. 
For each tile, the least-shadowed acquisition is selected as a weak reference, and all other acquisitions of that tile are paired with this reference to form shadowed–clean pairs. 
The DSM-based shadow masks and the DSMs themselves serve as auxiliary inputs, providing physically grounded cues about scene structure and illumination.

% The proposed pipeline is as follows: 
In the following, we describe the proposed pipeline consisting of
\begin{enumerate*}[label={\roman*)}, itemjoin={,\xspace}, itemjoin*={, and\xspace}, after={. }]
    \item standardise metadata
    \item sample fixed-size windows
    \item select a target date per window based on shadow content
    \item pair it with a geometrically and temporally compatible input date
    \item link elevation data
    \item record results in per-scene IDs
    \item produce train, validation, test splits from scenes with valid pairs
\end{enumerate*}

\begingroup
\setcounter{secnumdepth}{4}
\renewcommand{\theparagraph}{\roman{paragraph})} 

\paragraph{Data Organisation and Matching}
\label{sec:data_org}
Each scene is associated with RGB images, binary shadow masks, and acquisition metadata. 
A common index is created, linked to the image, masks, and metadata triplet, keeping only indices that are present in all required modalities. 
Acquisition timestamps from heterogeneous formats are converted to timezone-aware \emph{coordinated universal time (UTC)}. 
The scene footprint is summarised by a bounding polygon or bounding box, and key viewing and solar geometry (\ie off-nadir angle, look azimuth, sun elevation, sun azimuth) is derived for further downstream filtering.

\paragraph{Window Sampling}
\label{sec:window_sampling}
For each scene, a fixed number of $K$ random windows of size $\qtyproduct[parse-numbers=false]{CxC}{\pixel}$ (\eg $K=10, C=576$) is drawn using a reproducible seed. 
These windows define the crops used to evaluate local shadow content and to specify exact crop coordinates in the input-target data pair.

\paragraph{Per-window Target Selection and Pairing}
\label{sec:target_selection}
For each window $w \subset \R^2$ and acquisition date $t$, let $S_t(x,y)\in\{0,1\}$ be the binary S-EO shadow mask
($1 = \text{non-shadow}, 0 = \text{shadow}$). 
We measure the \emph{clear} (non-shadow) fraction% in $w$ as
\begin{align*}
\rho(w,t) &= \frac{1}{\abs{w}}\sum_{(x,y)\in w} S_t(x,y).
\end{align*}
in $w$.
The target date
\begin{align*}
    t^\star &= \operatornamewithlimits{arg\,max}_t \rho(w,t)
             = \operatornamewithlimits{arg\,min}_t \left( 1-\rho(w,t) \right),
\end{align*}
is chosen as the least-shadowed (\ie most clear) observation,
which is equivalent to minimising the shadow fraction, since it corresponds to $1-\rho(w,t)$.

\paragraph{Data-quality Constraints} 
\label{sec:data_quality}
Among dates $i \neq t^\star$, a single input time step is selected if it satisfies temporal, geometric, and data-quality constraints (\cf Table \ref{tab:thresholds}) and exhibits sufficient shadow contrast, \ie
\begin{align*}
    \Delta s(w; t^\star,i) &= \big| s(w,t^\star) - s(w,i) \big| \ge \tau_{\text{shadow}}\;.
\end{align*}
At most one input is kept per target per window to avoid redundancy. 
In this study,  $\tau_{\text{shadow}} = 0.1$ is selected. 
Furthermore, a temporal filtering is applied. 
A season-aware wrap-around gap is enforced through
\begin{align*}
    \Delta_d(t^\star,i) = \min\!\bigl(\lvert d(t^\star)-d(i)\rvert,\; 365-\lvert d(t^\star)-d(i)\rvert\bigr) \le \tau_{\text{season}}
\end{align*}
with $d(i)$ denoting the be day-of-year.
A stricter limit $\tau_{\text{winter}}$ may be applied for winter months November -- February %by substituting $\tau_{\text{season}}$ with $\tau_{\text{winter}}$. 
% The rationale is 
to limit the scene changes due to tree canopies and vegetation.

To further ensure high data quality in the target shadow-free reference, a candidate $i$ is accepted only if all of the following conditions hold (\cf \zcref{tab:thresholds}):
% \begin{itemize}[label={--}, noitemsep]
% \label{Section:_data_quality}
%   \item \textbf{Footprint overlap:} The Intersection-over-Union between target and candidate footprints exceeds $\tau_{\text{IoU}}$.
%   \item \textbf{View-geometry proximity:} The differences in off-nadir angle, look azimuth, and sun elevation are below $\tau_{\text{off}}, \tau_{\text{az}}, \tau_{\text{sun}}$, respectively. Circular angle differences use
%   \begin{align*}
%   \Delta\theta(\theta_1,\theta_2) &= \pi - \left| \pi - \left| \theta_1 - \theta_2 \right| \right|.
%   \end{align*}
%   \item \textbf{No-data check:} The fraction of no-data pixels in the candidate RGB crop does not exceed $\tau_{\text{nodata}}$.
% \end{itemize}
\begin{description}[noitemsep]
\label{Section:_data_quality}
  \item[Footprint overlap:] 
  The \emph{intersection-over-union} between target and candidate footprints exceeds $\tau_{\text{IoU}}$.
  \item[View-geometry proximity:] 
  The differences in off-nadir angle, look azimuth, and sun elevation are below $\tau_{\text{off}}, \tau_{\text{az}}, \tau_{\text{sun}}$, respectively. 
  For circular angle differences, we use
  \begin{align*}
  \Delta\theta(\theta_1,\theta_2) &= \PI - \abs{ \PI - \abs{ \theta_1 - \theta_2 } }.
  \end{align*}
  \item[No-data check:] 
  The fraction of no-data pixels in the candidate RGB crop does not exceed $\tau_{\text{nodata}}$.
\end{description}
If no candidate passes, the scene is skipped due to poor alignment with the reference acquisition.

\paragraph{Digital Surface Model Association}
\label{sec:dsm_ass}
If a DSM is available, its path is linked to each accepted pair, enabling downstream data processing to incorporate surface height information.

\paragraph{Data Pair and IDs Creation}
\label{sec:data_pair_ids}
For every accepted pair (input, target) and window, a YAML entry records the file paths to the input-target RGB and masks, the associated metadata paths, the crop coordinates $(y, x, h, w)$, the scene IDs, and the DSM file path. 
These IDs are used to generate data pairs during training/evaluation.

\paragraph{Dataset Splits}
\label{sec:dataset_splits}
A scene-level split is adopted to avoid spatial data leakage between the training, validation, and test sets. 
Each scene tile is assigned to exactly one split, so that no overlapping footprints or near-duplicate acquisitions appear across sets. 
The final configuration comprises \num{37} training scenes, \num{7} validation scenes, and \num{9} test scenes (\cf \zcref{tab:splits}). 
In particular, each city is represented in all three splits, and distinct scenes from every city are held out for validation and testing, enabling both within-city and cross-city generalisation to be assessed.

Splits are reported as scene-ID lists and recorded in YAML files in the accompanying code repository. 
All thresholds (\cf \zcref{tab:thresholds}), sampling hyperparameters $(K, C)$, file-naming patterns, and dataset paths are defined in a structured configuration, yielding a deterministic process given a fixed seed. 
The outcome consists of 
\begin{enumerate*}[label={\roman*)}, itemjoin={,\xspace}, itemjoin*={, and\xspace}, after={. }]
    \item paired, crop-level samples with strong inter-date shadow contrast and controlled geometric differences
    \item per-scene YAML IDs that reproduce every crop
    \item consistent scene-level train, validation, and test splits for fair evaluation
\end{enumerate*}
\endgroup

\begin{table}%[h]
  \caption{Dataset splits grouped by city. The number of samples refers to paired image crops per split.}
  \centering
  \scriptsize
  \setlength{\tabcolsep}{4pt}
  \begin{tabularx}{\columnwidth}{@{} l c c >{\raggedright\arraybackslash}X @{}}
    \toprule
    \textbf{Split} & \textbf{\# Scenes} & \textbf{\# Samples} & \textbf{Scene IDs (grouped by city)} \\
    \midrule
    Train & 37 & 255 &
      \textbf{UCSD}: 75, 721, 741, 744;
      \textbf{OMA}: 290, 385, 387, 478, 734, 766, 798, 831, 835, 927;
      \textbf{JAX}: 170, 203, 204, 275, 276, 295, 296, 313, 557, 584, 618, 652, 725, 726, 727, 728, 762, 763, 799, 800, 801, 802, 837 \\[2pt]
    Val   & 7  & 44 &
      \textbf{UCSD}: 742; 
      \textbf{OMA}: 291, 423, 799; 
      \textbf{JAX}: 172, 230, 764 \\[2pt]
    Test  & 9  & 63 &
      \textbf{UCSD}: 189; 
      \textbf{OMA}: 760; 
      \textbf{JAX}: 171, 231, 254, 255, 314, 530, 560 \\
    \bottomrule
  \end{tabularx}
  \label{tab:splits}
\end{table}

\subsection{Geometric Harmonisation}
A practical challenge when working with the S-EO dataset is the prevalence of off-nadir acquisitions. 
Differences in viewing geometry are typical in satellite imagery and can weaken alignment between paired dates. 
The effect is most pronounced in dense urban areas, where occlusions and parallax distort the image projection despite unchanged scene geometry.
Our pipeline addresses this by using RPC metadata to align and normalise geometry across dates and viewing directions in sensor space through the use of feature-based aligners. 
Pairs that remain outside predefined geometric tolerances after harmonisation are discarded, since they do not contribute to good reconstructions for the model. 
This treatment allows the use of a broader range of imagery, including opportunistic acquisitions that would otherwise be difficult to employ. 
It also supports applications where controlled acquisition is limited, such as disaster response or resource-constrained settings, and may extend to contexts with weak calibration. 
Pair selection proceeds in two passes:
\begin{description}
    \item[Pretraining:] broad thresholds (\cf \enquote{pretraining} column in \zcref{tab:thresholds}) to maximise coverage; residual viewpoint differences are handled by feature-based alignment (\cf \zcref{Section:alignement_feature_based}).
    \item[Refinement:] tighter thresholds (\cf \enquote{refined} column in \zcref{tab:thresholds}) for comparisons with methods assuming strong alignment (\eg UAV-based pipelines).
\end{description}
We regulate seasonal proximity with $\tau_{\text{season}}$ using day-of-year, so cross-year matches are allowed within the same tolerance, \eg {\DTMlangsetup{abbr}\DTMdate{2015-11-22}} \vs {\DTMlangsetup{abbr}\DTMdate{2016-11-18}}.
Because our dataset is restricted to the northern hemisphere, thresholds $\tau_{\text{winter}}$ are further tightened from November to March to prevent leaf-off/leaf-on canopy changes and snow coverage that would otherwise promote blurred outputs.

\begin{table}[t]
  \caption{Thresholds for constructing noisy–ground-truth pairs. The \enquote{Refined} column indicates any overrides or additional constraints imposed by the refined filtering procedure.}
  \centering
  \scriptsize
  \setlength{\tabcolsep}{4pt}
  \begin{tabularx}{\columnwidth}{@{} l c c >{\raggedright\arraybackslash}X @{}}
    \toprule
    \textbf{Parameter} & \textbf{Pretraining} & \textbf{Refined} & \textbf{Explanation} \\
    \midrule
    $\tau_{\text{IoU}}$        & $0.7$        & $0.7$        & Minimum bounding box overlap \\
    $\tau_{\text{off}}$      & $90^\circ$    & $7^\circ$    & Maximum allowed difference in camera viewing angle \\
    $\tau_{\text{az}}$        & $100^\circ$   & $10^\circ$   & Maximum allowed difference in azimuth angle \\
    $\tau_{\text{sun}}$  & $90^\circ$  & $90^\circ$  & Maximum allowed difference in sun elevation angle \\
    $\tau_{\text{nodata}}$         & $0.03$       & $0.03$       & Maximum allowed fraction of missing pixels \\
    $\tau_{\text{season}}$          & $60$ days    & $30$ days    & Maximum allowed difference between acquisition seasonal drift \\
    $\tau_{\text{winter}}$ & $60$ days          & $10$ days    & Tighter limit applied during winter period\\
    $\tau_{\text{shadow}}$   & $0$          & $0.1$        & Minimum fraction of difference between shadowy pixels \\
    \bottomrule
  \end{tabularx}
  \label{tab:thresholds}
\end{table}

\zcref[S]{fig:example_filter-az,fig:example_filter-off-nadir,fig:example_filter-temp} 
show representative rejections across three filters to visualise the effect of our geometric and temporal pairing constraints---\ie
azimuth difference, 
off-nadir angle, and 
seasonal/temporal proximity, respectively---%
in accordance with the corresponding thresholds summarised in \zcref{tab:thresholds}. 
These filters, tailored to the S-EO multi-date, multi-angle setting, are the basis of our deSEO pairing pipeline and prevent geometrically or radiometrically inconsistent matches from entering training.

\begin{figure}[t]
  \centering
  \begin{subfigure}[b]{0.32\linewidth}
    \includegraphics[width=\linewidth]{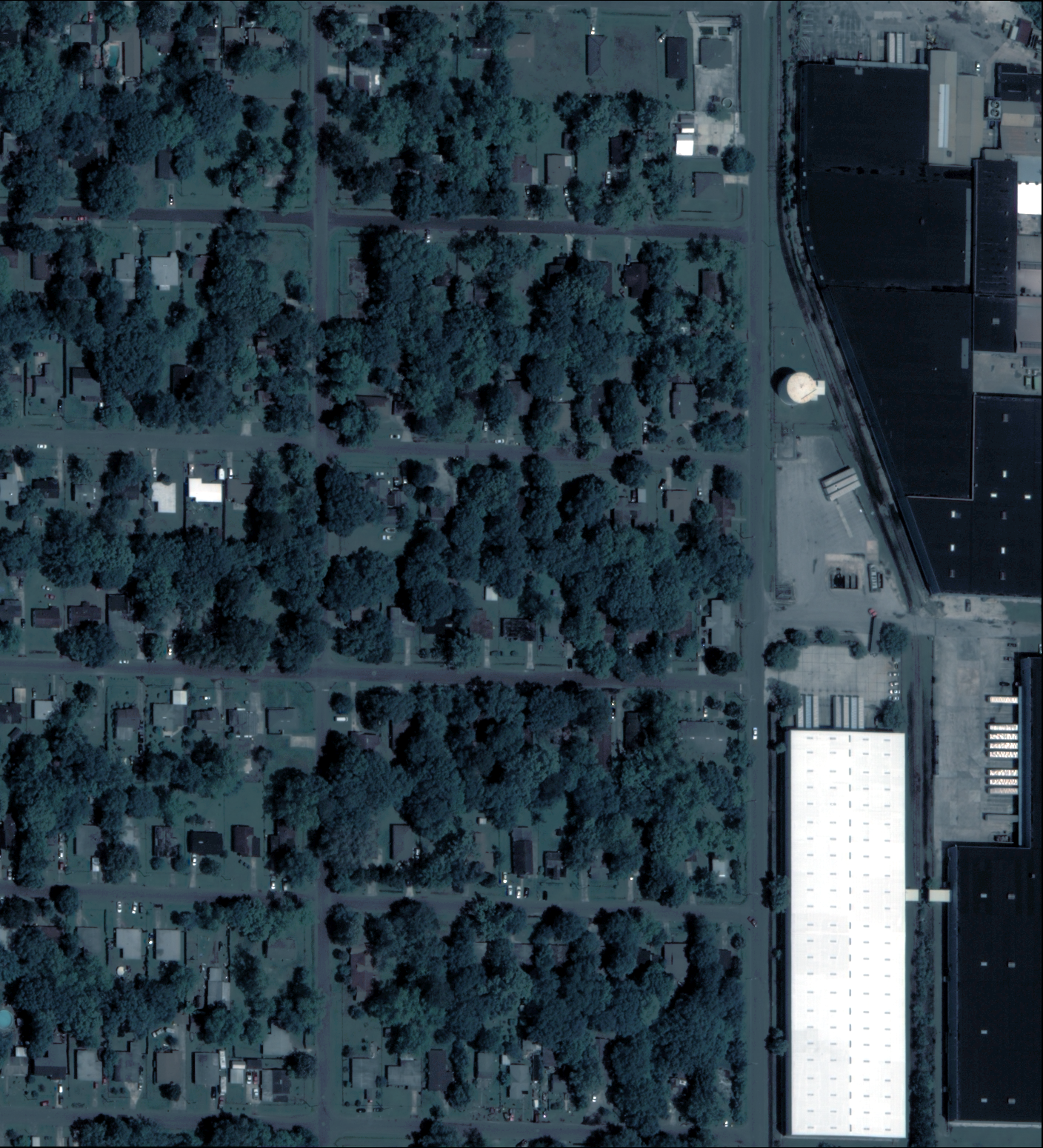}%
  \end{subfigure}\hfill
  \begin{subfigure}[b]{0.32\linewidth}
    \includegraphics[width=\linewidth]{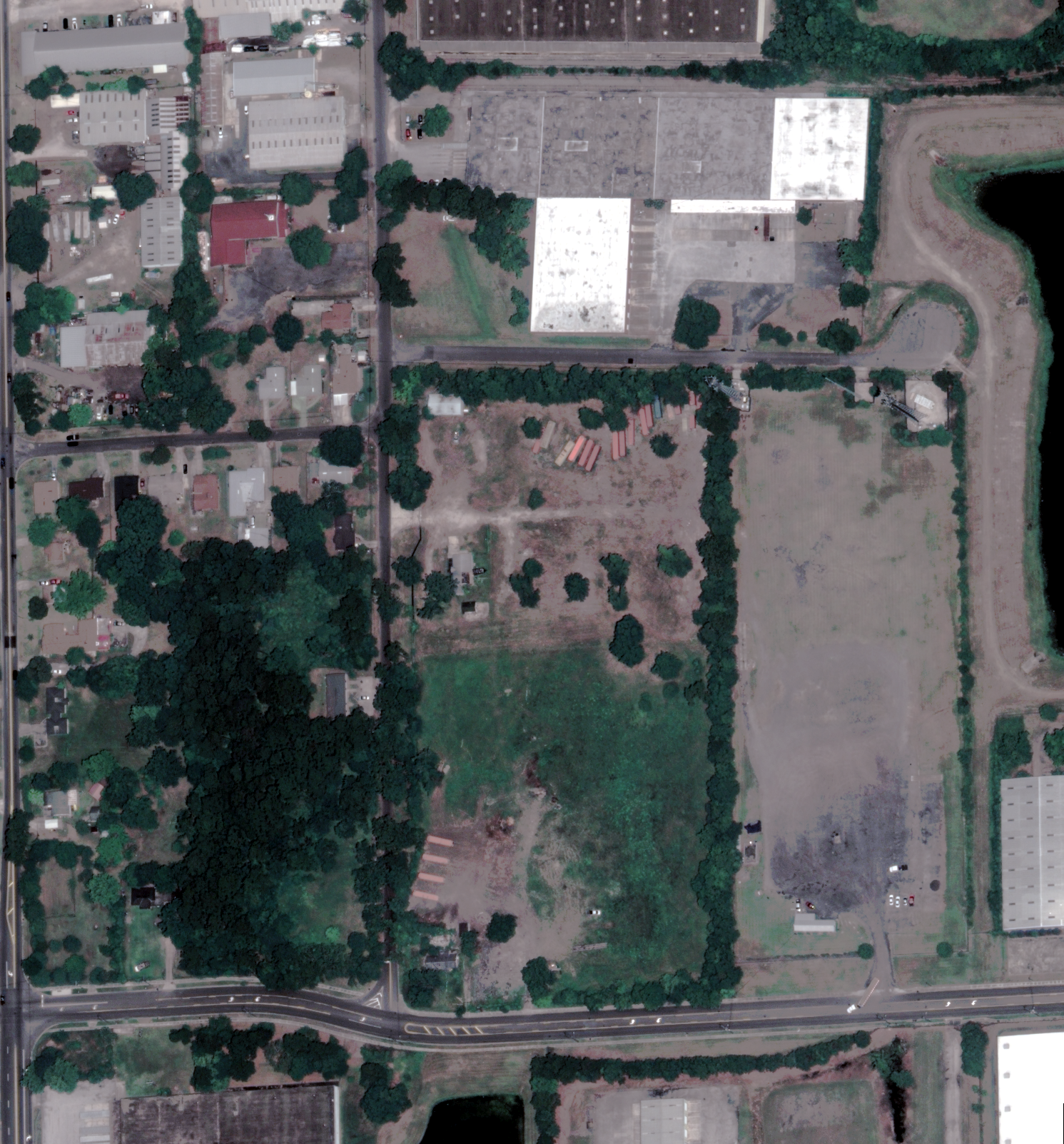}%
  \end{subfigure}\hfill
  \begin{subfigure}[b]{0.32\linewidth}
    \includegraphics[width=\linewidth]{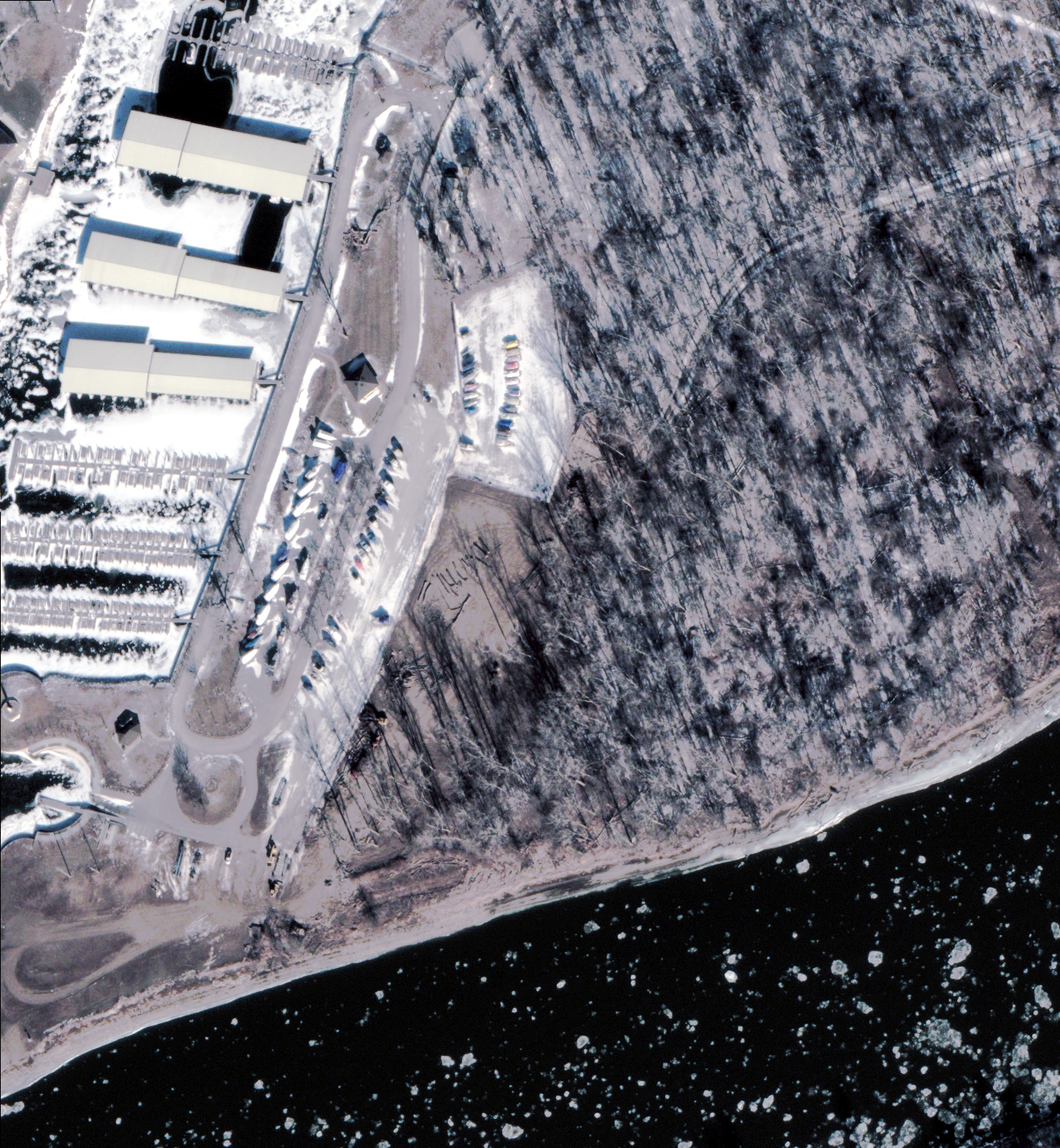}%
  \end{subfigure}\\
  \subcaptionbox{\label{fig:example_filter-az}}[.32\linewidth]{%
    \includegraphics[width=\linewidth]{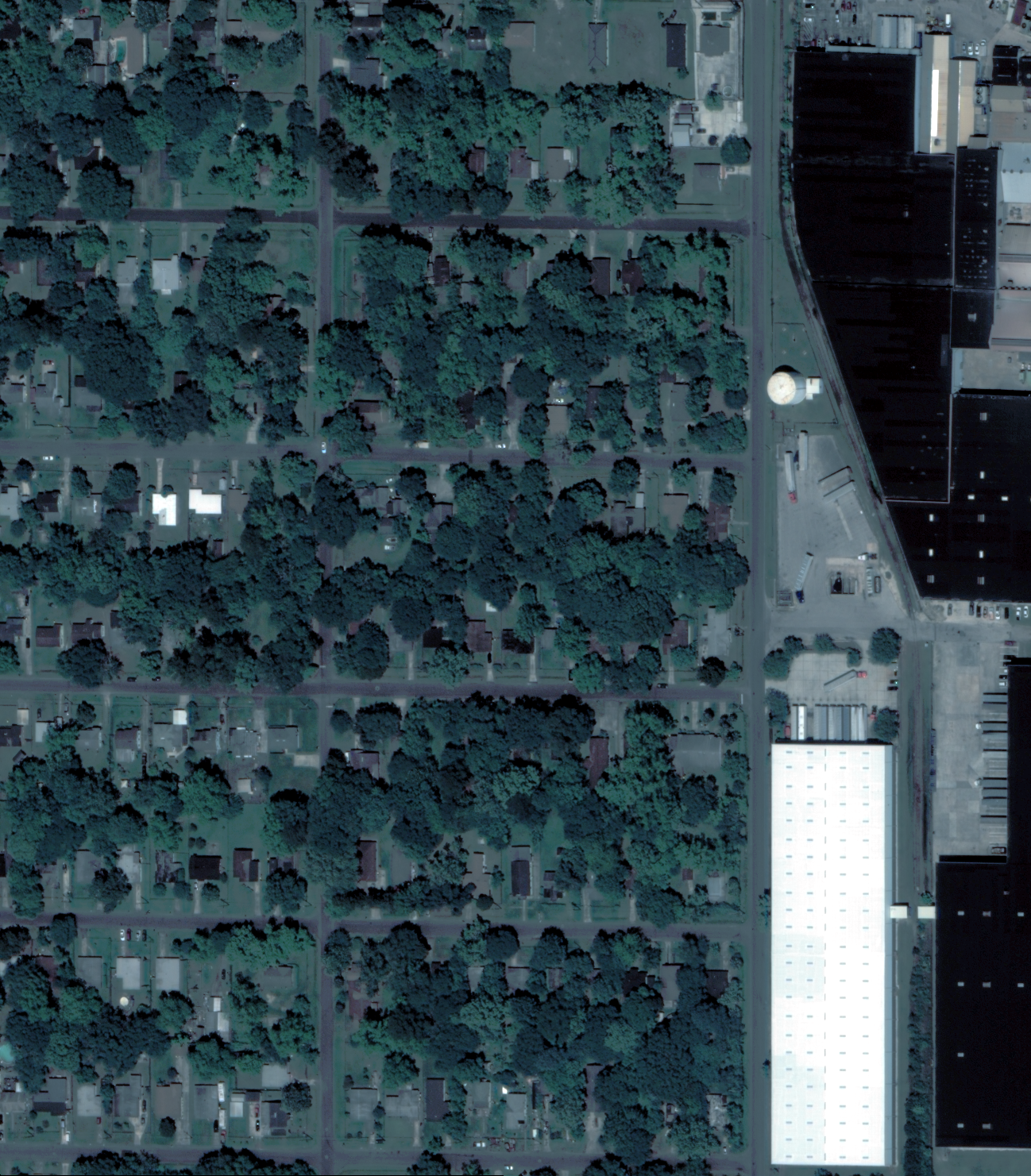}%
  }\hfill
  \subcaptionbox{\label{fig:example_filter-off-nadir}}[.32\linewidth]{%
    \includegraphics[width=\linewidth]{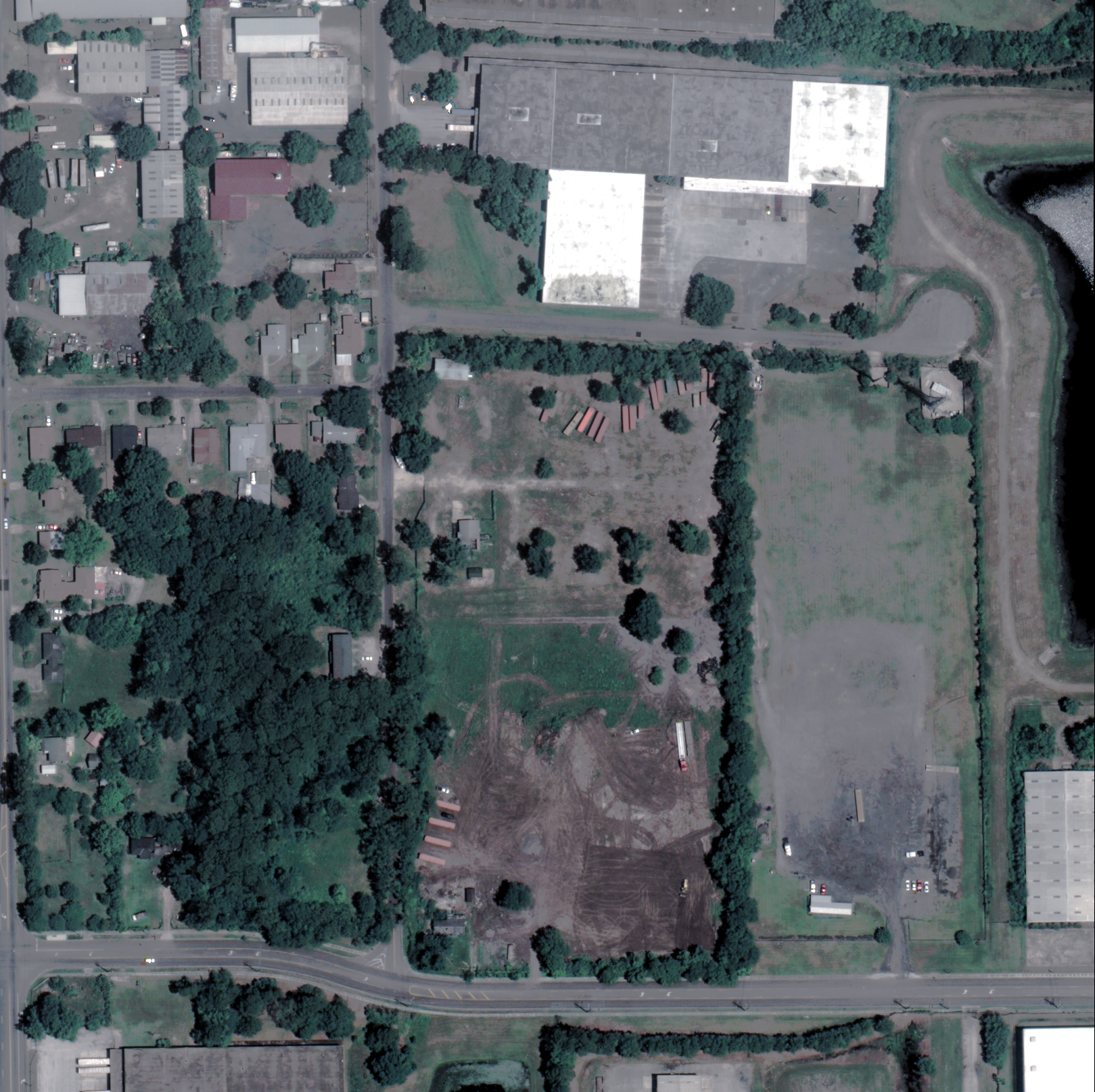}%
  }\hfill
  \subcaptionbox{\label{fig:example_filter-temp}}[.32\linewidth]{%
    \includegraphics[width=\linewidth]{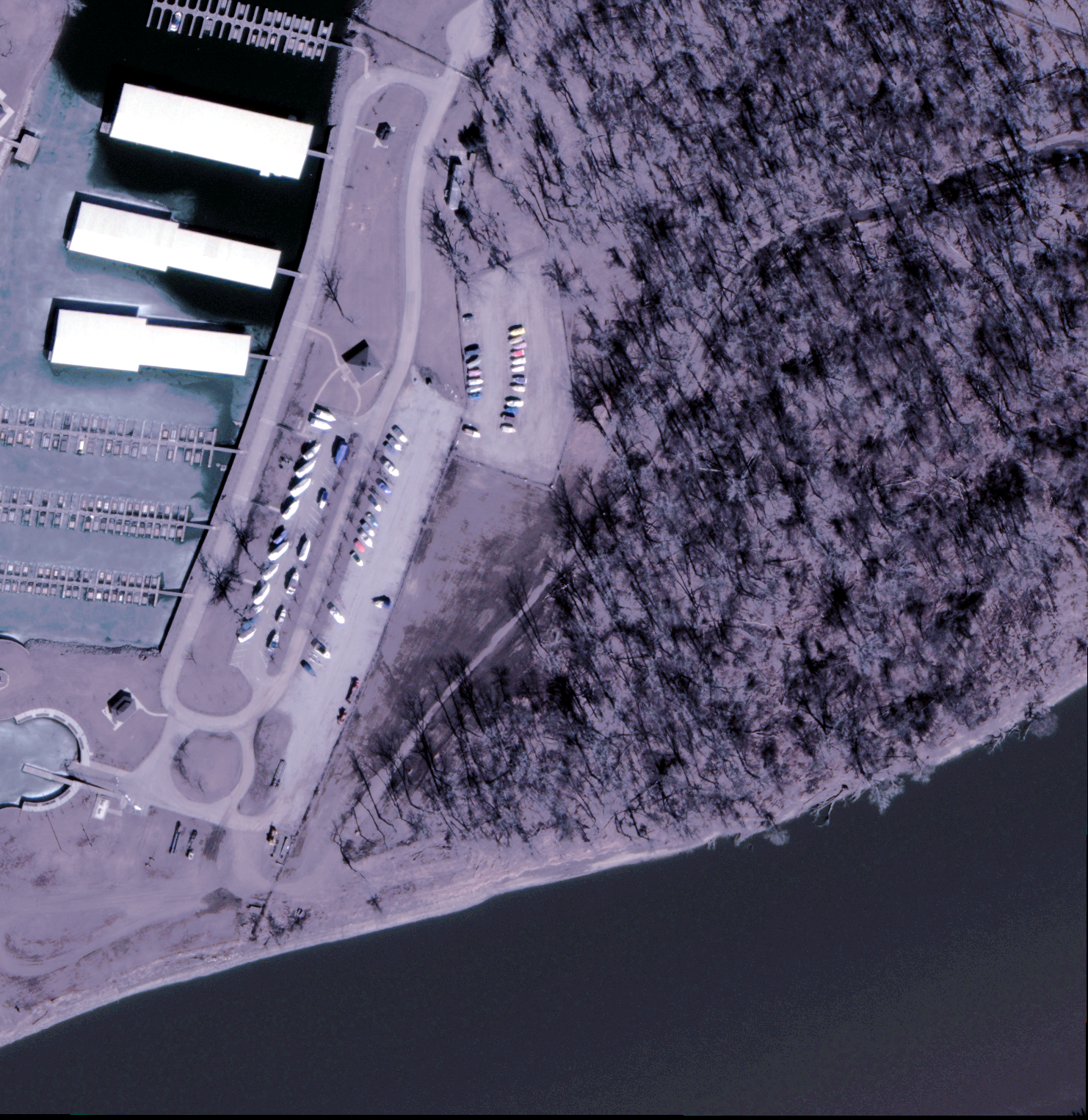}%
  }%
\caption{Rejected examples for 
  \textup{(a)} azimuth, 
  \textup{(b)} off-nadir, and 
  \textup{(c)} seasonal/temporal constraints following the corresponding threshold in \zcref{tab:thresholds}.}
  \label{fig:example_filter}
\end{figure}

\subsection{Orientation Normalisation}
\label{sec:Orientation_normalisation}
Additionally, due to variations in WorldView 30 satellite orbits, georeferenced imagery often differs in axis orientation. 
To correct for this, we approximate the Jacobian
\begin{align}
    J : (\text{lon}, \text{lat}) &\mapsto (\text{row}, \text{col})
    &=
    \begin{bmatrix}
        \dfrac{\partial \,\text{row}}{\partial \,\text{lon}} & 
        \dfrac{\partial \,\text{row}}{\partial \,\text{lat}} \\[1.2ex]
        \dfrac{\partial \,\text{col}}{\partial \,\text{lon}} & 
        \dfrac{\partial \,\text{col}}{\partial \,\text{lat}}
    \end{bmatrix},
\label{eq:jacobian}
\end{align}
of the mapping 
from world coordinates $(\text{lon}, \text{lat})$ to image coordinates $(\text{row}, \text{col})$ which we estimate using central finite differences. 
The image orientation is determined by the relative magnitudes of derivatives (to detect axis swaps) and their signs (to detect flips/rotations). The cases are summarised in \zcref{tab:jacobian-cases}.

\begin{table}[t]
  \centering
  \caption{%
    Orientation diagnosis from Jacobian derivatives (\zcref{eq:jacobian}) and their corrections. 
    % CCW and CW stand for counterclockwise and clockwise rotations. H. and V. stand respectively for Horizontal and Vertical.
  }
  \label{tab:jacobian-cases}
  \scriptsize
  \setlength{\tabcolsep}{3.5pt}
  \begin{tabularx}{\columnwidth}{@{} c c c c c >{\raggedright\arraybackslash}X @{}}
    \toprule
    Case & $\partial \mathrm{row}/\partial \mathrm{lon}$ & $\partial \mathrm{row}/\partial \mathrm{lat}$ & $\partial \mathrm{col}/\partial \mathrm{lon}$ & $\partial \mathrm{col}/\partial \mathrm{lat}$ & Orientation \\
    \midrule
    \multirow{2}{*}{(A)} 
      & $>0$ & ---   & ---     & $>0$ & $\qty{-90}{\degree}$ \\
      &      & otherwise &     &      & $\qty{90}{\degree}$ \\
    \midrule
    \multirow{4}{*}{(B)} 
      & ---   & $<0$ & $>0$ & ---   & correct \\
      & ---   & $>0$ & $>0$ & ---   & V. flip \\
      & ---   & $<0$ & $<0$ & ---   & H. flip \\
      & ---   & $>0$ & $<0$ & ---   & $\qty{180}{\degree}$ rotation \\
    \bottomrule
  \end{tabularx}

  % \vspace{2pt}
  \scriptsize
  (A) $\lvert\partial \mathrm{row}/\partial \mathrm{lon}\rvert \gg \lvert\partial \mathrm{row}/\partial \mathrm{lat}\rvert$, \;
  $\lvert\partial \mathrm{col}/\partial \mathrm{lat}\rvert \gg \lvert\partial \mathrm{col}/\partial \mathrm{lon}\rvert$\\
  (B) $\lvert\partial \mathrm{row}/\partial \mathrm{lat}\rvert \ge \lvert\partial \mathrm{row}/\partial \mathrm{lon}\rvert$, \;
  $\lvert\partial \mathrm{col}/\partial \mathrm{lon}\rvert \ge \lvert\partial \mathrm{col}/\partial \mathrm{lat}\rvert$\\
  % CW: clock-wise, CCW: counter-clock-wise, 
  H: horizontally, V: vertically
  \vspace*{\baselineskip}
\end{table}

\begin{figure}[t]
  \centering
  \hspace*{0.10\linewidth}%
  \subcaptionbox{before preprocessing\label{fig:orientation_normalisation-before}}[.28\linewidth]{\includegraphics[width=\linewidth]{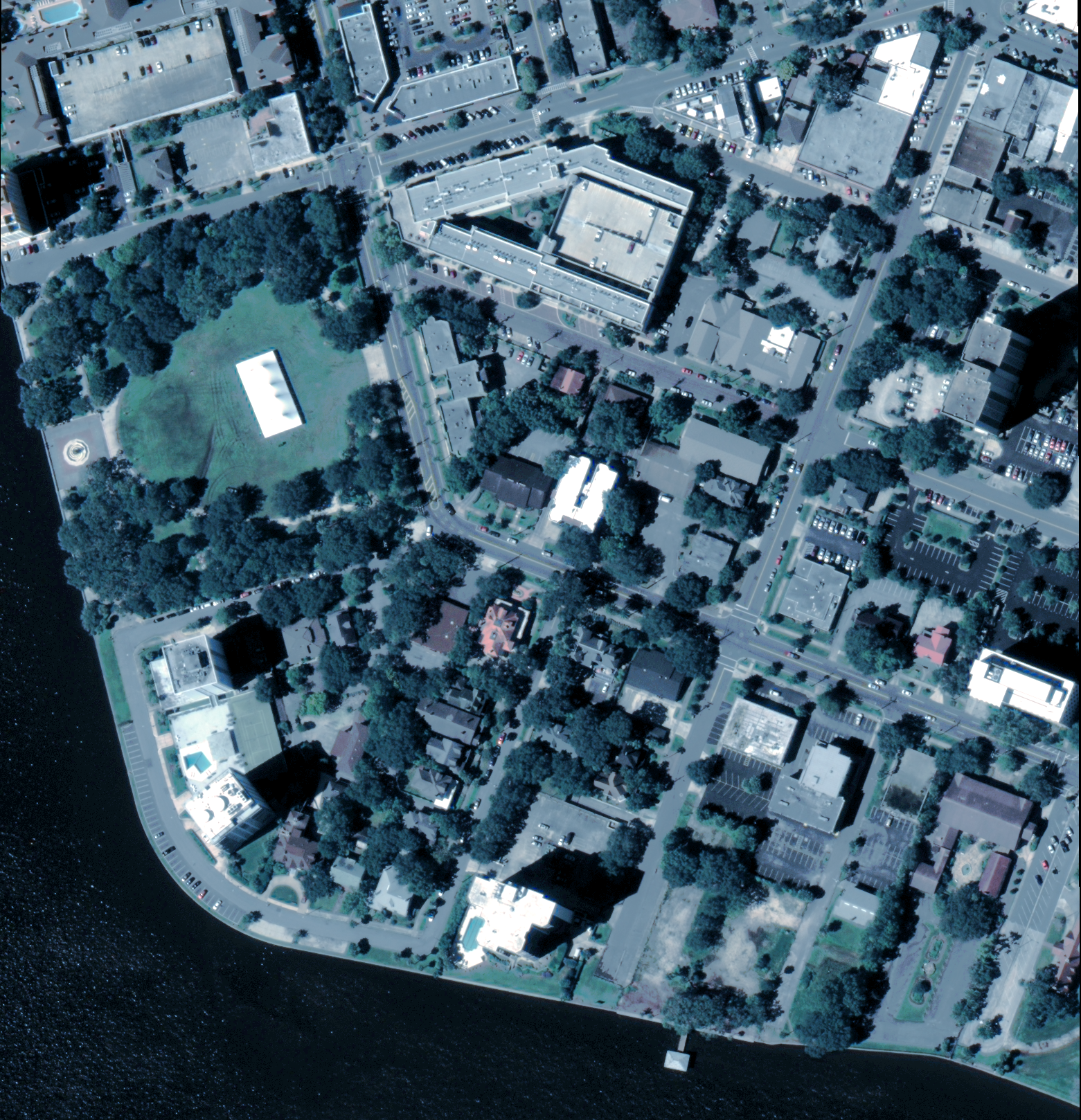}}\hfill
  \hspace{0.04\linewidth}%
  \subcaptionbox{after preprocessing\label{fig:orientation_normalisation-after}}[.28\linewidth]{\includegraphics[width=\linewidth]{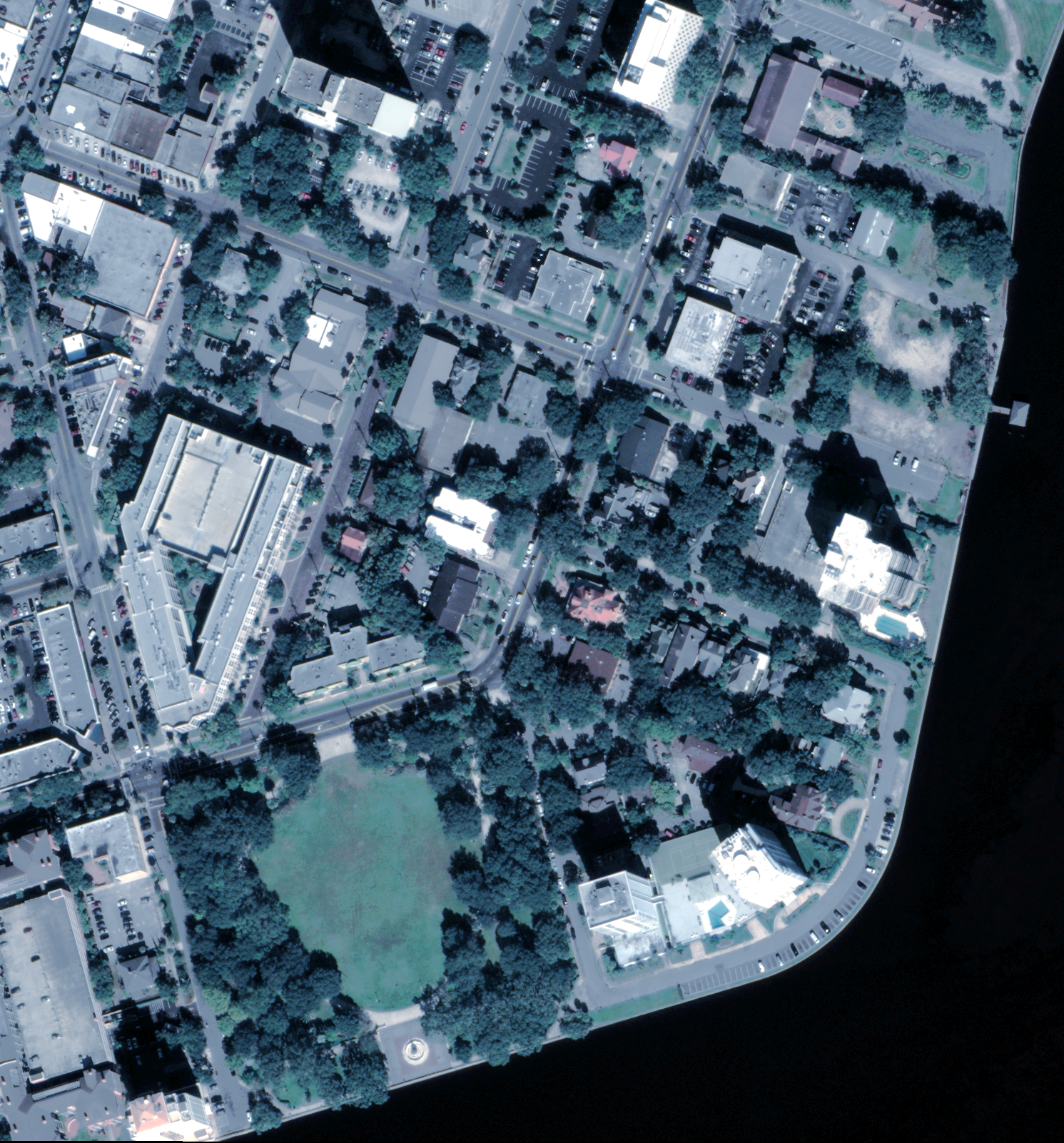}}
  \hspace*{0.10\linewidth}%
\caption{%
    Examples for 
    \textup{(a)} the unprocessed acquisition with incorrect rotation and
    \textup{(b)} the same area after orientation correction during preprocessing, as described in \zcref{tab:jacobian-cases}.%
}
  \label{fig:orientation_normalisation}
\end{figure}

\subsection{Pairwise Registration and Validity Mask}
\label{Section:alignement_feature_based}
Since a residual perspective distortion persists in the acquired data due to the $\qty{10}{\degree}$ tolerance in the off-nadir angle, we perform a feature-based registration between shadowy and non-shadowy image pairs. 
Preliminary experiments with lightweight descriptors such as \emph{oriented FAST and rotated BRIEF (ORB)} \citep{rublee2011orb} and \emph{accelerated features (XFeat)} \citep{potje2024xfeat} yield insufficient alignment quality. 
We therefore adopt the detector-free \emph{local feature transformer (LoFTR)} framework \citep{sun2021loftr}, which has demonstrated strong performance in remote sensing contexts \citep{jovhari2023performance}. 
In our pipeline, grayscale input and target images are processed with the outdoor-pretrained LoFTR model, using a confidence threshold of $0.20$ and a cap of \num{8000} correspondences, to obtain dense keypoint matches. 
The images input into LoFTR are downscaled to have a maximum width of $W=\qty{1024}{\pixel}$ and height $H$ scaled accordingly. 
Affine transformations are then estimated bidirectionally using \emph{random sample consensus (RANSAC)} \citep{fischler1981random}, and the model with the highest inlier count and lowest median reprojection error is selected. 
Finally, to ensure supervision is applied only where cross-view correspondence is well-defined, we propagate a binary validity mask $M\in\{0,1\}^{H\times W}$ alongside each training pair after warping. 
The $L_1$ loss of the \emph{pix2pix} generator and the discriminator inputs are restricted to the valid subset by masking and renormalisation. 

\subsection{Neural Network Architecture}
We frame shadow removal as a conditional image-to-image translation task driven by multi-date, multi-geometry supervision. 
The model consists of a U-Net generator \parencite{Ronneberger15:CNB} that predicts a residual deshadowing correction and a PatchGAN-style discriminator \parencite{isola2017image} enhanced by a soft, mask-driven attention prior. 
Together, they leverage geometry-aware cues and validity masking to learn illumination-consistent corrections while preserving radiometry outside shadow regions.

In all the following runs, optimisation uses Adam~\citep{kingma2014adam}, a cosine learning-rate schedule~\citep{loshchilov2016sgdr} over 100 epochs, and weights initialisations drawn from a normal distribution with a gain of $0.02$. 
Training is performed with a batch size of $\numproduct{576x576}$ tiles, using random crops of size $\qtyproduct{256x256}{\pixel}$, and employs two discriminator updates per generator update during the first 20~epochs.
Applying self-attention at intermediate feature resolutions best preserves fine-grained structure in shadow removal.
This placement sharpens edges and micro-textures while avoiding over-smoothing at high resolutions and detail loss at coarse scales.
It balances local detail with contextual consistency, preserving high-frequency content.
By contrast, using attention only in the generator weakens adversarial supervision and produces overly smooth results.
Ablation results support this design choice (cf. \zcref{tab:ablation_val}).

\subsubsection{Generator}
The proposed model follows a pix2pix-style \emph{conditional generative adversarial network (cGAN)} architecture \parencite{isola2017image} in which the generator $G$ is realised by a U-Net with skip connections. 
Unlike \emph{embedding space consistency networks \parencite[ESCNet; ][]{luo2023evolutionary}}, where shadow removal using a \emph{style-guided re-deshadow network \parencite[SRNet; ][]{Wan22:SGS}} and radiometric correction through a \emph{radiation adjustment network (RANet)} are separated, we employ a single end-to-end translation model. 
This unified formulation reflects the deSEO setup, where multi-date, multi-geometry satellite supervision and DSM-derived priors favour a geometry-aware treatment. 
Optional spectral normalisation \parencite{miyato2018spectral} and non-local self-attention layers, following the \emph{self-attention generative adversarial network (SAGAN)} design \parencite{zhang2019self}, allow each spatial location to attend globally for improved context modelling. 
When available, DSMs are concatenated with RGB inputs to form a 4-channel condition for both $G$ and $D$. 

All reconstruction and perceptual objectives are computed only on valid RGB pixels, excluding DSM and padded regions according to the masks introduced in \zcref{Section:alignement_feature_based}. 
The reconstruction term can be configured as $L_1$, SSIM, VGG-perceptual~\parencite{johnson2016perceptual}, or LPIPS with VGG backbone~\citep{zhang2018unreasonable}
\begin{align*}
    \mathcal{L}_{\text{rec}}^{L_1}(\hat{B}, B) 
    &= \lVert \hat{B} - B \rVert_1,\\
    \mathcal{L}_{\text{rec}}^{\text{VGG}}(\hat{B}_{\text{rgb}}, B_{\text{rgb}})   
    &= \sum_{\ell}w_\ell\lVert\phi_\ell(\hat{B}_{\text{rgb}})-\phi_\ell(B_{\text{rgb}})\rVert_{1}, \text{or} \\
    \mathcal{L}_{\text{rec}}^{\text{LPIPS}}(\hat{B}_{\text{rgb}}, B_{\text{rgb}})
    &= \text{LPIPS}_{\text{VGG}}(\hat{B}_{\text{rgb}},B_{\text{rgb}}),
\end{align*}%
respectively.
Perceptual losses are evaluated on RGB tensors normalised to ImageNet mean and standard deviation, and are preferred to mitigate blur under imperfect alignment. 
To maintain colour and hue stability in non-shadowed regions, we apply a colour-consistency loss and an HSV-based regularization
\begin{align*}
    \mathcal{L}_{\text{color}} &= \|(\hat{B}_{\text{rgb}}-A_{\text{rgb}})\odot M_\text{tar}\|_{1} \quad \text{and}\\
    \mathcal{L}_{\text{$L_1$ HS}} &= \|[\Delta H,|S_{\hat{B}}-S_A|]\odot M_\text{tar}\|_{1} \quad,
    % \text{with}\quad \Delta H &= \min(|H_{\hat{B}}-H_A|,1-|H_{\hat{B}}-H_A|) \quad,
\end{align*}
respectively,
with $\Delta H = \min(|H_{\hat{B}}-H_A|,1-|H_{\hat{B}}-H_A|)$.

In addition to the mask-restricted objectives, we include a lightly weighted global $L_1$ loss term $\mathcal{L}_{{+L_1}} $ to stabilise optimisation. 
This provides a uniform gradient across all valid pixels, especially in weakly aligned regions and prevents global colour drift without overriding the perceptual and colour-consistency losses. 

The full generator objective 
\begin{align*}
\mathcal{L}_G &= \mathcal{L}_{\text{GAN}}
               + \lambda_{\text{rec}}\mathcal{L}_{\text{rec}}
               + \lambda_{\text{col}}\mathcal{L}_{\text{color}}
               + \lambda_{L_1\,\text{HS}}\mathcal{L}_{\text{HS}} \notag \\
               &\phantom{=} + \lambda_{+L_1}\mathcal{L}_{{+L_1}} 
\end{align*}
combines all components
and is applied only on valid correspondences to ensure stability under weak alignment. 
Optionally, inputs or gradients corresponding to pixels that remain shadowed in both views can be zeroed out using the prior mask $S$.

\subsubsection{Discriminator}
To capture high-frequency details, we focus the discriminator on local image neighbourhoods rather than the entire image, following the classical PatchGAN approach \citep{isola2017image}. 
The discriminator classifies whether each overlapping $N\times N$ patch appears real or generated, operating fully convolutional across the image. 
The resulting patch-wise responses are then aggregated (\eg averaged) to produce the final output of the discriminator $D$.

We enhance this design with a \emph{soft shadow attention mask (SSAM)} that focuses the adversarial signal on regions transitioning from shadow to non-shadow. 
Given binary input and target masks $M_\text{in},M_\text{tar} \in \{0,1\}^{W \times H}$, respectively, the attention prior
\begin{align*}
  \operatorname{SSAM}(x,y) &= (1-M_\text{in}(x,y))M_\text{tar}(x,y),
\end{align*}
emphasises areas of shadow disappearance. 
For each layer $l$, the feature map% $\phi_\ell$% is modulated by a normalised attention map
\begin{align*}
    \phi_l &\leftarrow \phi_l \odot (1+\gamma_l\widehat{S}_l) 
    && \text{with} &
    \widehat{S}_l &= \tfrac{S_l}{\mathrm{mean}(S_l)+\varepsilon}
\end{align*}
is modulated by a normalised attention map,
where $\gamma_l$ is a learnable gating parameter. 
The optional suppression of gradients from fully shadowed pixels can be enforced via the relative flag. 
All convolutional layers use spectral normalisation and a dropout layer with $p=0.1$ follows each downsampling block. 
Training supports vanilla, \emph{least-squares generative adversarial network \parencite[LSGAN; ][]{Mao17:LSG}}, or hinge adversarial losses, with their optional relativistic variants. 
One-sided label smoothing (\eg $y_{\text{real}} \in [0.9,1]$) and an additional penalty 
$
\lambda_{\!R_1} \cdot \mathbb{E}\|\nabla_x D(x)\|_2^2
$
on real samples
are available for regularisation. Early training may use $k_D>1$ discriminator updates per generator step to stabilise the adversarial game. 
This geometry-aware discriminator emphasises shadow transitions while preserving radiometric stability elsewhere, complementing  $G$'s reconstruction and perceptual losses.

\section{Results}

\begin{table*}[t]
  \centering
  \scriptsize
  \caption{
  Ablation study on the validation set (mean $\pm$ standard deviation).
  Standard deviations rounded to one significant digit; means adjusted accordingly.
  Metrics computed on RGB channels.
  }
  \label{tab:ablation_val}
  \setlength{\tabcolsep}{6pt}
  \begin{tabular}{@{}lcccccc@{}}
    \toprule
    \textbf{Configuration} &
    \textbf{PSNR $\uparrow$} &
    \textbf{RMSE $\downarrow$} &
    \textbf{SSIM $\uparrow$} &
    \textbf{$L_1$ $\downarrow$} &
    \textbf{VGG19 $L_1$ $\downarrow$} &
    \textbf{LPIPS $\downarrow$} \\
    \midrule

    \textbf{Baseline} &
    $\mathit{18 \pm 2}$ &
    $32 \pm 5$ &
    $\mathbf{0.6 \pm 0.1}$ &
    $\mathbf{0.09 \pm 0.01}$ &
    $\mathbf{0.44 \pm 0.04}$ &
    $\mathbf{0.41 \pm 0.07}$ \\

    \textbf{RGB only (remove DSM)} &
    $9 \pm 2$ &
    $90 \pm 20$ &
    $0.18 \pm 0.06$ &
    $0.30 \pm 0.05$ &
    $0.57 \pm 0.06$ &
    $0.71 \pm 0.02$ \\

    \textbf{No G self-attention} &
    $17 \pm 2$ &
    $37 \pm 8$ &
    $0.5 \pm 0.1$ &
    $0.10 \pm 0.02$ &
    $0.47 \pm 0.05$ &
    $0.42 \pm 0.07$ \\

    \textbf{Gamma features disabled} &
    $17 \pm 2$ &
    $35 \pm 9$ &
    $\mathbf{0.6 \pm 0.1}$ &
    $0.10 \pm 0.03$ &
    $0.46 \pm 0.04$ &
    $0.42 \pm 0.06$ \\

    \textbf{No HV regularisation} &
    $16 \pm 2$ &
    $39 \pm 9$ &
    $0.4 \pm 0.1$ &
    $0.11 \pm 0.03$ &
    $0.46 \pm 0.05$ &
    $0.45 \pm 0.08$ \\

    \textbf{$L_1$-only (no perceptual)} &
    $\mathbf{18 \pm 1}$ &
    $\mathbf{31 \pm 5}$ &
    $\mathbf{0.6 \pm 0.1}$ &
    $\mathbf{0.09 \pm 0.01}$ &
    $0.55 \pm 0.05$ &
    $0.47 \pm 0.04$ \\

    \textbf{Reduced latent channels} &
    $17 \pm 2$ &
    $35 \pm 7$ &
    $0.5 \pm 0.1$ &
    $0.10 \pm 0.01$ &
    $0.46 \pm 0.04$ &
    $0.44 \pm 0.06$ \\

    \textbf{No spectral normalisation} &
    $16 \pm 2$ &
    $39 \pm 7$ &
    $0.5 \pm 0.1$ &
    $0.12 \pm 0.02$ &
    $0.49 \pm 0.05$ &
    $0.46 \pm 0.07$ \\

    \textbf{Shadow attention disabled} &
    $\mathbf{18 \pm 1}$ &
    $34 \pm 3$ &
    $\mathbf{0.6 \pm 0.1}$ &
    $0.10 \pm 0.01$ &
    $0.45 \pm 0.05$ &
    $0.42 \pm 0.06$ \\

    \textbf{No Pretraining} &
    $17 \pm 2$ &
    $36 \pm 8$ &
    $0.5 \pm 0.1$ &
    $0.10 \pm 0.01$ &
    $0.46 \pm 0.05$ &
    $0.42 \pm 0.06$ \\

    \textbf{No dropout} &
    $17 \pm 2$ &
    $37 \pm 7$ &
    $\mathbf{0.6 \pm 0.1}$ &
    $0.10 \pm 0.02$ &
    $0.47 \pm 0.05$ &
    $0.42 \pm 0.06$ \\    
    
    \bottomrule
  \end{tabular}
\end{table*}

Before developing our architecture, we trained the original SRNet of \textcite{luo2023evolutionary} on deSEO-generated paired data.
Training quickly became unstable, with non-convergent adversarial loss and severe generator artefacts.
We attribute this to SRNet’s reliance on tightly aligned UAV imagery with near-pixel correspondence, assumptions violated by our multi-temporal, multi-geometry satellite pairs, which contain residual misalignment and scene changes.
We therefore use SRNet as the closest transfer baseline.
Its failure indicates that UAV-oriented shadow-removal models do not readily transfer to high-resolution satellite data, motivating the geometry-aware design of deSEO and our model.
This negative result motivated the development of a dedicated architecture specifically designed for high-resolution satellite imagery. 
In particular, we adopt a geometry-aware formulation that emphasises perceptual and feature-space losses, rather than purely pixel-wise objectives, in order to improve robustness under imperfect correspondences and illumination variability. 
The following experiments evaluate this model on the deSEO dataset, analyse the impact of key design choices, and quantify the contribution of geometry-aware inputs and perceptual supervision.

\begin{figure}[t]
  \centering
  % \setlength{\tabcolsep}{1pt}
  % \renewcommand{\arraystretch}{0}

  % \begin{tabular}{ccc}
  %   \includegraphics[width=.32\linewidth]{figures/pretraining_fake_B.jpeg} &
  %   \includegraphics[width=.32\linewidth]{figures/pretraining_real_A.jpeg} &
  %   \includegraphics[width=.32\linewidth]{figures/pretraining_real_B.jpeg} \\
  %   \scriptsize (a) Reconstruction &
  %   \scriptsize (b) Input &
  %   \scriptsize (c) Target \\
  % \end{tabular}
  \subcaptionbox{Reconstruction}[.32\linewidth]{\includegraphics[width=\linewidth]{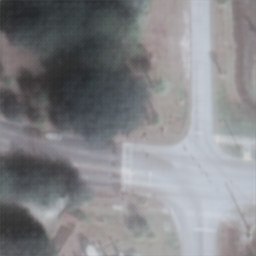}}\hfill
  \subcaptionbox{Input}[.32\linewidth]{\includegraphics[width=\linewidth]{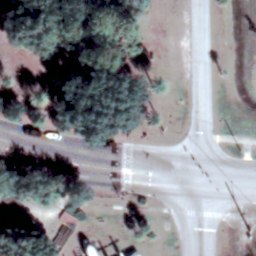}}\hfill
  \subcaptionbox{Target}[.32\linewidth]{\includegraphics[width=\linewidth]{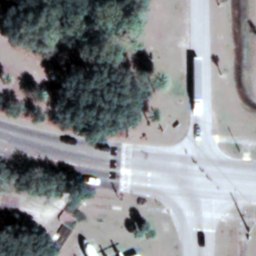}}
  \caption{Examples from the validation set during pretraining.}
  \label{fig:val-three-tiles_pretraining}
\end{figure}

We first explore a range of training configurations to identify models that can reconstruct high-resolution satellite imagery and learn useful features for shadow removal.
Using the pretraining dataset obtained with the filtering procedure in \zcref{tab:thresholds}, this stage does not yet produce high-quality reconstructions but encourages the generator to learn structural and radiometric priors of satellite imagery.
Representative validation examples are shown in \zcref{fig:val-three-tiles_pretraining}.

Pretraining prioritises optimisation stability over deshadowing quality.
Low, balanced learning rates and a mild discriminator warm-up prevent early divergence, while the loss emphasises $L_1$ and HVS terms, disabling colour, perceptual, and attention losses to reduce adversarial sensitivity before meaningful spatial structure is learned.
DSM input and relatively large model capacities are retained to enable early geometry-aware feature learning.

Although pretraining alone does not yield accurate reconstructions or deshadowing, it establishes structural and radiometric priors.
We then fine-tune on the refined dataset (\cf \zcref{tab:thresholds}), containing higher-quality, more consistent pairs.
The core optimisation setup is kept fixed, varying only capacity and regularisation.
Finetuning uses longer training, smaller batches, spectral normalisation in the discriminator, dropout in the generator, and activation of geometry-aware modules (Gamma and shadow attention) with DSM input.
These changes stabilise adversarial training and exploit geometric cues while limiting overfitting.

We validate these design choices through an ablation study (\zcref{tab:ablation_val}).
The full model achieves the best balance across metrics, with the highest SSIM and the lowest VGG19 and LPIPS.
Removing DSM causes the largest degradation, halving PSNR and SSIM and confirming its central role.
Disabling self-attention or gamma features leads to moderate drops, mainly in structural fidelity and perceptual quality, while removing HV regularisation or spectral normalisation slightly increases reconstruction and perceptual errors.
Overall, DSM is the dominant factor, with spectral normalisation and HV regularisation having smaller individual effects.

\begin{figure}[t]
    \centering
    % \setlength{\tabcolsep}{2pt}
    % \begin{tabular}{cccc}
    %     \includegraphics[width=0.24\linewidth]{comparison_real_A_00000000.jpeg} &
    %     \includegraphics[width=0.24\linewidth]{comparison_real_B_00000000.jpeg} &
    %     \includegraphics[width=0.24\linewidth]{baseline_fake_B.jpeg} &
    %     \includegraphics[width=0.24\linewidth]{l1_fake_B.jpeg} \\
    %     (a) Input & (b) Target & (c) Full model & (d) L1-only
    % \end{tabular}
    \subcaptionbox{input}[.24\linewidth]{\includegraphics[width=\linewidth]{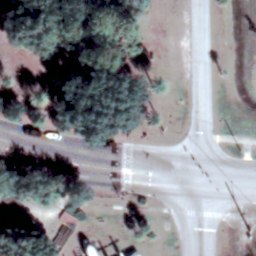}}\hfill
    \subcaptionbox{target}[.24\linewidth]{\includegraphics[width=\linewidth]{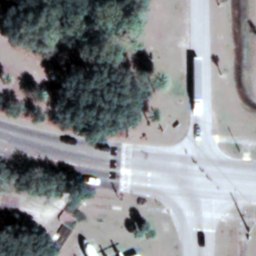}}\hfill
    \subcaptionbox{full model}[.24\linewidth]{\includegraphics[width=\linewidth]{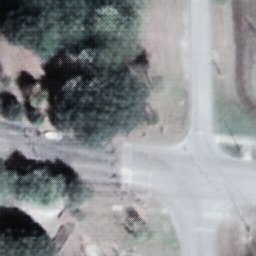}}\hfill
    \subcaptionbox{$L_1$-only}[.24\linewidth]{\includegraphics[width=\linewidth]{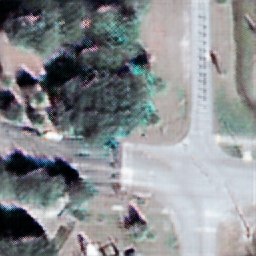}}
    \caption{
        Qualitative comparison of the full model and the $L_1$-only ablation.
    }
    \label{fig:l1_ablation}
\end{figure}

Some variants, such as the $L_1$-only model, achieve strong pixel-wise scores (highest PSNR/RMSE and low $L_1$; \zcref{tab:ablation_val}) but fail at deshadowing.
They prioritise intensity matching over shadow removal, yielding worse perceptual metrics and visibly retaining shadows (\cf \zcref{fig:l1_ablation}).
Reducing the channel capacity similarly induces a predictable decline, although the model remains relatively robust to such architectural compression. 
Overall, the results highlight that DSM input, perceptual supervision, and stabilising regularisation (HV, spectral norm, attention) contribute jointly to the model’s ability to preserve both radiometric and perceptual fidelity while performing meaningful shadow reduction.

\begin{table}[t]
  \centering
  \scriptsize
  \caption{Final metrics evaluated on the test set.}
  \label{tab:test_results}
  \begin{tabular}{l c}
    \toprule
    \textbf{Metric} & \textbf{Mean $\pm$ Std} \\
    \midrule
    PSNR (\unit{\deci\bel}) $\uparrow$                & $18 \pm 1$ \\
    RMSE $\downarrow$                   & $34 \pm 8$ \\
    SSIM $\uparrow$                     & $0.49 \pm 0.08$ \\
    $L_1\; [0,1]$ $\downarrow$         & $0.09 \pm 0.03$ \\
    Perceptual (VGG19 $L_1$) $\downarrow$  & $0.42 \pm 0.08$ \\
    LPIPS (VGG) $\downarrow$            & $0.46 \pm 0.05$ \\
    \bottomrule
  \end{tabular}
\end{table}

\begin{figure}[t]
  \centering
  % \setlength{\tabcolsep}{1pt}
  % \renewcommand{\arraystretch}{0}
  % \begin{tabular}{ccc}
  %   \includegraphics[width=.32\linewidth]{figures/test_fake_B.jpeg} &
  %   \includegraphics[width=.32\linewidth]{figures/test_real_A.jpeg} &
  %   \includegraphics[width=.32\linewidth]{figures/test_real_B.jpeg} \\
  %   \scriptsize (a) Reconstruction &
  %   \scriptsize (b) Input &
  %   \scriptsize (c) Target \\
  % \end{tabular}
  \subcaptionbox{reconstruction}[.32\linewidth]{\includegraphics[width=\linewidth]{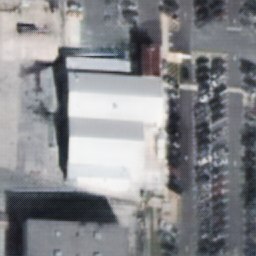}}\hfill
  \subcaptionbox{input}[.32\linewidth]{\includegraphics[width=\linewidth]{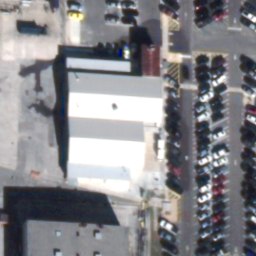}}\hfill
  \subcaptionbox{target}[.32\linewidth]{\includegraphics[width=\linewidth]{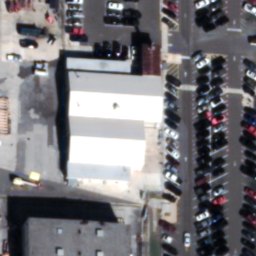}}
  \caption{Examples from the test set on the baseline model.}
  \label{fig:test-three-tiles}
\end{figure}

To fully assess the model's performance, we evaluate it on the test set, with results provided in \zcref{tab:test_results}. 
A qualitative example is presented in \zcref{fig:test-three-tiles}.
The final test run uses the same configuration as our baseline model (Exp.~4 in \zcref{tab:ablation_val}). 
The results on the held-out test set are presented in \zcref{tab:test_results}. 
Similarly to validation (\cf \zcref{tab:ablation_val}), the model favours perceptual fidelity (low VGG19-$L_1$ and LPIPS, competitive $L_1$) over distortion metrics such as PSNR and RMSE. 
This mirrors why the baseline configuration performed best on validation: it optimises SSIM, $L_1$, and perceptual metrics rather than maximising PSNR/RMSE (\zcref{tab:ablation_val}). 
Moreover, this behaviour is not confined to the validation split—the same pattern appears in the test set, suggesting that it is a stable property of the model rather than an observation on a particular subset. 
The relatively large standard deviations on test (\eg SSIM $\pm 0.08$, RMSE $\pm 8$) suggest scene-dependent variance, likely tied to shadow coverage and illumination. 
This spread is partially attributable to the limited size of the validation and test splits.
In future iterations, we plan to expand the dataset, both in scale and geographic diversity, to better quantify generalisation across a broader range of conditions.
Because shadows occupy a small fraction of each image, purely global metrics can underweight improvements in shadowed regions. 
We therefore complement metrics with a blind manual review focused on cast-shadow boundaries and relit regions. 
Despite limited texture recovery, the method consistently identifies and reduces shadows (\cf \zcref{fig:test-three-tiles}). 
Nevertheless, the major limitation remains that the difference between the shadow profiles is often not significant across the image% 
 % (\cf \zcref{fig:test-three-tiles})
, so the reconstructed image does not yield a completely shadow-free reference. 
One reason might be the lower cardinality of the dataset, potentially not enabling the model to learn general shadow features.
% COMMENT I disagree with the penultimate sentence. It also conflates an observation with a hypothesis. You can say that the major limit is that the reconstructed image does not yield a completely shadow-free image, and that this might be due to the low difference in shadow profiles in the training set (although this is highly debatable). But stating the the major issue is the low difference is the shadow profiles in the training set which leads to non-completely shadow-free images makes the assumption that your hypothesis is correct.
% 
Given the limited extent of most shadows, traditional metrics are more informative about fine-detail reconstruction than deshadowing itself, justifying our use of an additional visual assessment. 
The results are qualitatively encouraging: the model can reliably identify shadows in unseen images under different lighting conditions and off-nadir perspectives, providing a basis for further improvement.
% COMMENT I have the feeling that claiming a manual assessment makes the lack of visual results in the paper much more glaring than I thought when I answered your question earlier about putting examples to support the ablation. Here the visual assessment feels a bit unsupported sincce you have no figure to refer to.

\section{Conclusion}

We address the challenge of generating reliable paired data for deshadowing in high-resolution satellite imagery, where the lack of true shadow-free references and the variability of acquisition geometries make supervised learning particularly difficult.
We introduced deSEO, a geometry-aware preprocessing and training pipeline that transforms shadow detection datasets into weakly supervised deshadowing datasets by exploiting multi-date acquisitions under explicit geometric, temporal, and radiometric constraints.
The pipeline produces reproducible window-level pairings, registration-driven validity masks, and scene-level splits suitable for weakly supervised deshadowing.

Using S-EO as a case study, we demonstrate that the proposed pipeline can derive high-quality training pairs from a detection-oriented resource.
We also propose a geometry-aware deshadowing model, inspired by SRNet but redesigned for satellite imagery, that can learn meaningful shadow reduction despite residual misalignment.
While global metrics on the test set show only moderate improvements, they are consistent across scenes and align with qualitative inspection, which confirms that the model consistently reduces the visual impact of cast shadows under diverse illumination and viewing conditions.
The ablation study further highlights the importance of DSM inputs, perceptual supervision, and stabilising regularisation in achieving perceptually coherent deshadowing (\cf \zcref{tab:ablation_val}).

The current size of the S-EO-derived dataset remains a critical limiting factor, contributing to the variance observed across scenes and constraining the model's ability to learn generalisable shadow patterns.
While the present results are best characterised as shadow reduction rather than complete shadow removal, we believe that deSEO provides a practical path toward full shadow removal in high-resolution satellite imagery.
By enabling the systematic creation of weakly supervised training data from multi-temporal observations, deSEO establishes the data foundation needed for models to progressively learn stronger shadow compensation under diverse illumination, seasonal, and viewing conditions.
We therefore view this work not as an endpoint, but as a first step toward fully shadow-free reconstruction.

Future work will focus on applying deSEO to additional datasets spanning different sensors and geographic regions, enriching the diversity of paired samples, and incorporating more advanced radiometric normalisation, uncertainty modelling, and modelling strategies that better disentangle shadow effects from intrinsic surface appearance.
Improved temporal and geometric alignment will also be important for moving from shadow reduction towards more complete shadow removal.
Finally, given the inherent difficulty of obtaining shadow-free ground truth, evaluating the effect of deshadowing on downstream tasks such as classification or change detection remains an important direction for future work and would enable a broader assessment of its practical impact.

Overall, deSEO establishes a first reproducible, physics-aware dataset creation methodology for weakly supervised deshadowing in high-resolution satellite imagery.
This framework provides a foundation upon which more robust and generalisable deshadowing methods can be developed, particularly in complex observation conditions where shadows and illumination effects degrade image quality.

\section*{Acknowledgments}
This work was carried out within the SAFIR research project funded by the Austrian Research Promotion Agency (FFG) as part of the Research, Technology \& Innovation (RTI) initiative \enquote{Digitaler Zwilling Österreich}. Code and materials are available on GitHub\footnote{\url{https://github.com/AIT-Assistive-Autonomous-Systems/deSEO}}.

\printbibliography

\end{document}